\documentclass[10pt,twocolumn,letterpaper]{article}

\usepackage{iccv}
\usepackage{times}
\usepackage{epsfig}
\usepackage{graphicx}
\usepackage{amsmath}
\usepackage{amssymb}
\usepackage{enumitem}
\usepackage{bbding} 
\usepackage{times}
\usepackage{epsfig}
\usepackage{graphicx}
\usepackage{amsmath}
\usepackage{amssymb}
\usepackage{array}

\usepackage{enumitem}

\usepackage{times}
\usepackage{epsfig}
\usepackage{graphicx}
\usepackage{amsmath}
\usepackage{amssymb}

\usepackage{enumitem}

\usepackage{booktabs}
\usepackage{multirow}
\usepackage{makecell}

\usepackage[colorlinks]{hyperref}
\usepackage{breakurl}

\usepackage{color, colortbl}
\definecolor{iGray}{gray}{0.9}
\definecolor{beaublue}{rgb}{0.74, 0.83, 0.9}
\definecolor{Royal_Blue}{rgb}{0.0, 0.1, 0.66}

\usepackage{pifont}
\newcommand{\xmark}{\ding{55}}%
\usepackage{bbding}

\usepackage{graphicx}
\usepackage{mwe}
\usepackage{amsmath, bm}
\usepackage{float}
\usepackage{caption}
\usepackage{pifont}
\usepackage{bbding}

\usepackage{graphicx}
\usepackage{mwe}
\usepackage{amsmath, bm}
\usepackage{float}
\usepackage{caption}

\usepackage{graphicx}
\usepackage{amsfonts}
\usepackage{color}
\usepackage{bm}
\usepackage{amsmath}
\usepackage{amssymb}
\usepackage{multirow}
\usepackage{makecell}
\usepackage{enumitem}
\usepackage{tabularx}
\usepackage{fixltx2e} 
\usepackage{booktabs} 
\usepackage{xcolor}
\usepackage{url}
\usepackage{graphicx}
\usepackage{amsfonts}
\usepackage{color}
\usepackage{bm}
\usepackage[accsupp]{axessibility}  

\iccvfinalcopy 

\begin{document}

\title{Learning Tracking Representations via Dual-Branch Fully Transformer Networks}

\author{Fei~Xie$^{1}$\textsuperscript{\thanks {Interns at MSRA} }, Chunyu Wang$^{2}$, Guangting Wang$^{2}$, Wankou Yang$^{1}$,  Wenjun Zeng$^{2}$\\
	$^{1}$Southeast University, China\\
	$^{2}$Microsoft Research Asia\\
	
	{\tt\small jaffe03@seu.edu.cn, chnuwa@microsoft.com}\\ 
	
	\tt\small{
		flylight@mail.ustc.edu.cn, wkyang@seu.edu.cn, wezeng@microsoft.com}
	
}

\maketitle

\begin{abstract}
   We present a Siamese-like Dual-branch network based on solely Transformers for tracking.  Given a template and a search image, we divide them into non-overlapping patches and extract a feature vector for each patch based on its matching results with others within an attention window.  For each token, we estimate whether it contains the target object and the corresponding size. The advantage of the approach is that the \textbf{features are learned from matching, and ultimately, for matching}. So the features are aligned with the object tracking task. The method achieves better or comparable results as the best-performing methods which first use CNN to extract features and then use Transformer to fuse them. It outperforms the state-of-the-art methods on the GOT-10k and VOT2020 benchmarks. In addition, the method achieves real-time inference speed (about $40$ fps) on one GPU. The code and models are released at \url{https://github.com/phiphiphi31/DualTFR}.
\end{abstract}


\section{Introduction}
Visual Object Tracking (VOT) is a fundamental problem in computer vision which aims at tracking an object of interest in a video given its bounding box in the first frame~\cite{OTB2013}. This is generally addressed by looking for the location in the search image whose features have the largest correlation with those in the template image. Introducing of deep Convolutional Neural Network (CNN) has notably boosted the tracking accuracy because of the improved features for matching ~\cite{SiameseFC,UPDT}. 

\begin{figure}[t]
	\centering{\includegraphics[scale = 0.28]{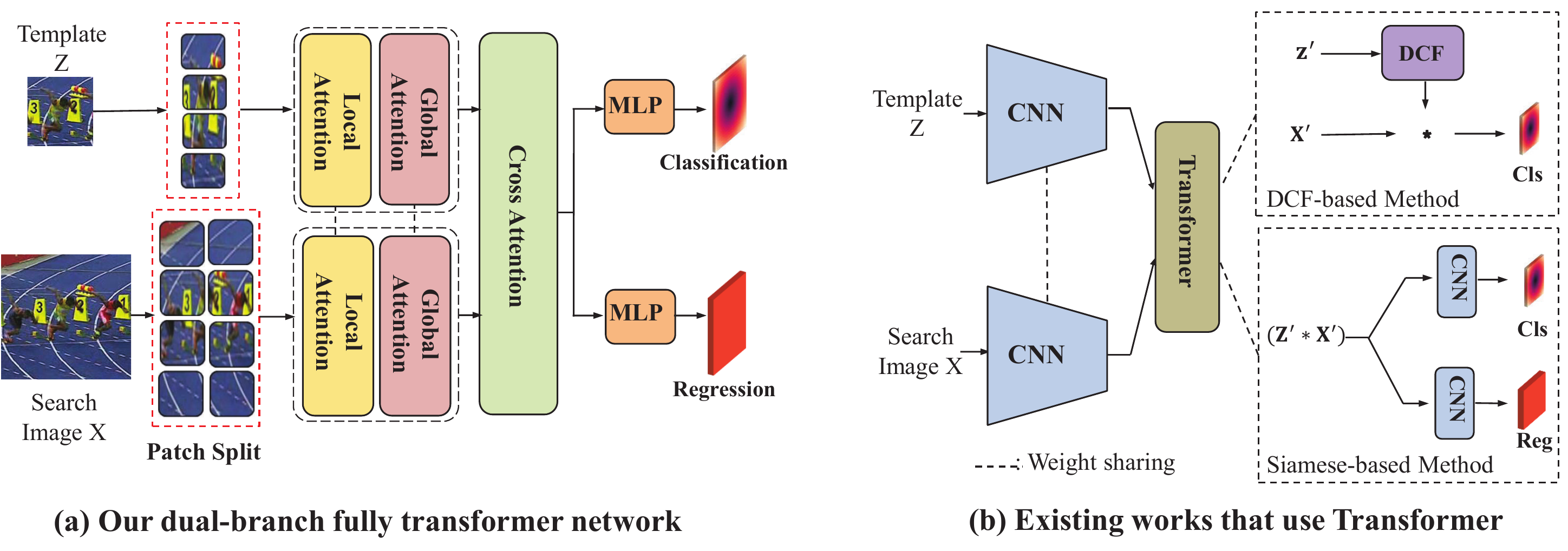}}
	\caption{Comparison of our full Transformer method (a) and the existing ``CNN+Transformer'' based methods (b) which first use CNN to extract features and then fuse them with Transformer networks. Siamese-based~\cite{SiameseFC} and DCF-based~\cite{DiMP, atom} methods are two popular pipelines in tracking. }
	\vspace{-3mm}
	\label{fig:compare}
\end{figure}

The core of VOT is to extract features that are not only robust to appearance variation of the same object in different frames, but also discriminative among different objects. To achieve the target, most of the recent tracking methods~\cite{siamrpn++,siamdw, transt, siamgat} manually select ``optimal'' features from either shallow or deep layers of CNN, or their fusion based on empirical experience \cite{he2018twofold}. But our experiments show that the features computed in this way are not optimal mainly because CNN is not specifically designed for the matching purpose. Instead, it only looks for the presence of certain image features of the interested classes but does not understand the structural dependency among regions of different objects in the image.

Vision Transformer (ViT)~\cite{vit}, which divides an image into regular tokens,
adds positional embedding, and encodes each token based on token-to-token similarities, is a promising approach to extract features for visual object tracking because it is aware of the dependency among all tokens (objects) in all encoding layers. In other words, it extracts features from matching, and for matching, which is consistent with the ultimate task.

Some recent works have already applied Transformer to VOT~\cite{Trackformer, tmt,transt,stark}. But most of them still heavily rely on CNN to extract features and only use Transformer in the last layer to fuse them by global attention. Although they have dramatically boosted the tracking accuracy on benchmark datasets, a natural question arises--- can Transformer also benefit the earlier feature extraction step since it can model the structural dependency among different regions? We aim to answer this question in this work.

In this work, we present the first study of using pure Transformers to extract features for tracking. To that end, a Siamese-like dual-branch network is proposed as shown in Fig.~\ref{fig:compare} (a). It divides the template and search images into tokens and extracts a feature for each based on its matching results with others in the same image. This is achieved by mixed efficient local attention blocks and powerful global attention blocks as shown in Fig.~\ref{fig:tfr}. In addition, we propose \emph{cross attention blocks} which fuse the tokens between the template and search image. This helps to learn features which are robust to variation in videos. We do not use cross attention in every layer because it is expensive and not necessary--- the template and search images are usually from neighboring frames thus having similar patterns which can be captured by the  local and global attention blocks. We find using a single cross attention block at the final layer is sufficient in our experiments.

To achieve a good balance between accuracy and speed, we use local attention model on the high-resolution feature maps of most shallow layers, and global attention model only on the low-resolution feature maps as shown in Fig. \ref{fig:tfr}. This notably improves the inference speed (about 40fps on a single 2080Ti GPU). On top of the computed features for each token, we add a MLP layer to estimate whether it is the target (classification head) and the size of the target box in current frame (regression head). Without bells and whistles, this simple approach already outperforms the state-of-the-art methods on multiple tracking benchmarks. We provide extensive ablation studies to validate different factors of the approach. In particular, we find that the use of transformer to extract tracking features is critical to the success of the approach. The main contributions of this work are summarized as follows:

\begin{itemize}
	\item We present the first attempt to use pure transformer network to compute tracking features, which according to our experiments, is superior to the dominant ``CNN+Transformer'' based methods.
	\item We introduce a very simple dual-branch architecture which consists of local attention blocks and global attention blocks in each branch, respectively, and cross attention block to fuse features between the template and search image. The approach achieves a good accuracy/speed trade-off. 
	\item The proposed approach outperforms the state-of-the-art methods, including the ``CNN+Transformer'' based methods, on multiple tracking benchmarks. In addition, we provide a lot of empirical experience to researcher/engineers in this field with extensive ablation studies.
	
\end{itemize}

\section{Related Work}

\subsection{Visual Object Tracking}
We classify the state-of-the-art object trackers into two classes. The first class is the Siamese-based methods which generally consist of three steps: CNN-based feature extraction for the template and search images, feature correlation, and a prediction head. For example, SiamFC~\cite{SiameseFC}, which is the pioneer work of the series of Siamese methods, directly locates the target object at the position with the largest correlation. SiamFC obtains promising results but it cannot estimate the size of the bounding box. SiamRPN applies a proposal network~\cite{li2018high,FasterRCNN} to the correlation map to find the object location and size which is more powerful than SiamFC. In addition, many works are introduced to improve Siamese trackers such as Feature Pyramid Network~\cite{siamrpn++, fpn}, deeper backbone~\cite{siamdw}, anchor-free detection ~\cite{SiamFCpp, SiamCAR} and feature-alignment~\cite{Ocean}.

The second class of methods are DCF-based~\cite{KCF, ECO, DiMP, PrDiMP, SuperDiMP, DCFST} which utilize online DCF to classify the target. A response map is generated by computing the correlation between the online DCF and the features in the search region. Current DCF methods are also heavily dependent on CNN-based deep features~\cite{UPDT, atom} and linear correlation filter.
Our approach belongs to the first class. But different from the previous works, we use fully Transformer networks to extract features.

\subsection{Vision Transformer}
The success of transformer in natural language processing has drawn wide attention from the computer vision community. 
The main advantage of ViT ~\cite{vit} over CNN is that the global dependency can be easily captured.
A variety of ViTs~\cite{dvit, godeeper, swin, twins, token} have been proposed which achieve state-of-the-art performance on many downstream computer vision tasks, such as object detection, semantic segmentation and human keypoint detection. 
Among them, DeepVit~\cite{godeeper} attempts to make the ViT structure go deeper for more powerful representations. 
PVT~\cite{pvt} adopts pyramid structure like CNNs to better adapt ViT to image tasks. %
SwinT~\cite{swin} restricts the self-attention operation within a local window which avoids quadratic complexity.
CrossVit~\cite{crossvit} proposes a dual-path transformer-based structure to extract multi-scale features for enhanced visual representations.

\begin{figure*}[t]
	\centering{\includegraphics[scale = 0.55]{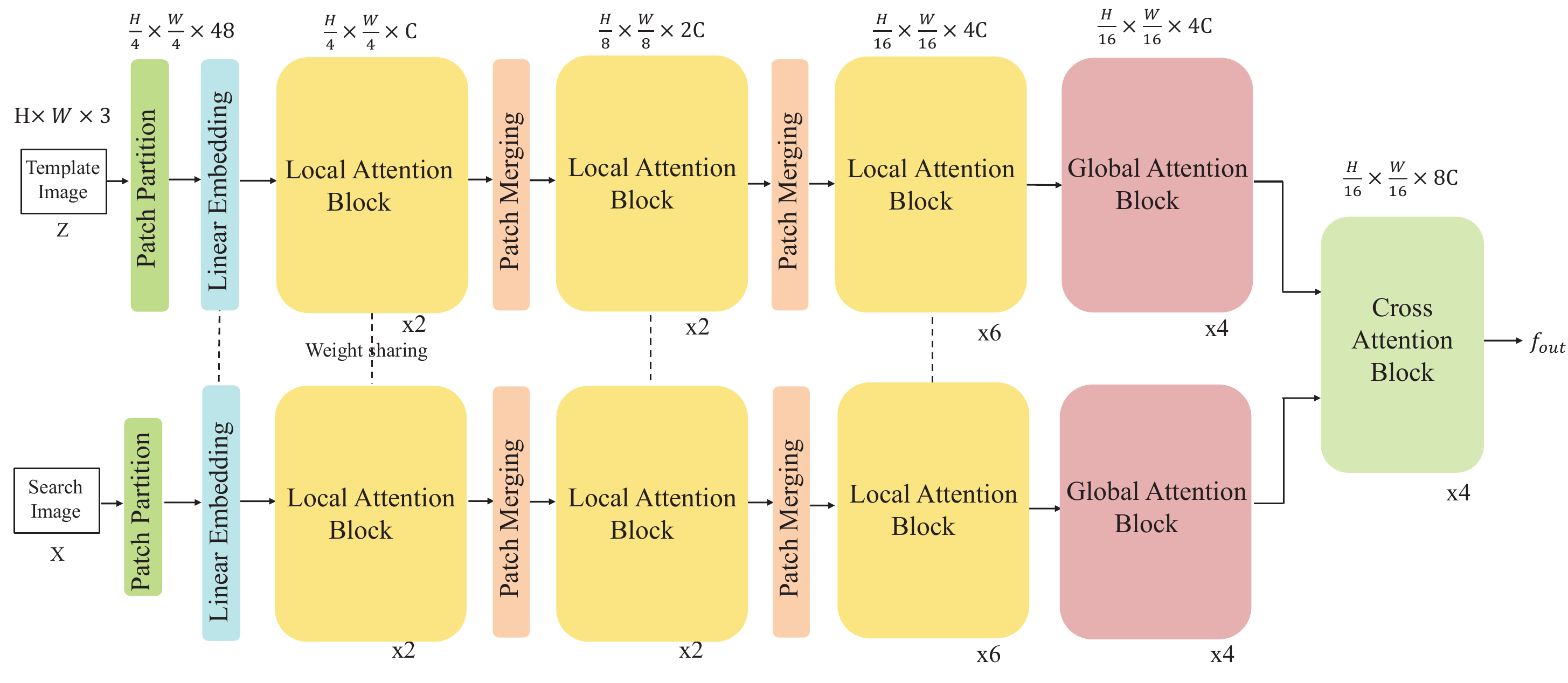}}
	\caption{The architecture of our dual-branch fully Transformer-based pipeline (DualTFR).}
	\vspace{-3mm}
	\label{fig:tfr}
\end{figure*}

\subsection{Transformer in Object Tracking} 
Some recent works have already explored to use Transformer in VOT~\cite{tmt,transt,stark}. In general, they still use CNN to extract features for the template and search image and use Transformer to enhance the CNN features which is the main difference from our fully transformer-based method. For example, TransMTracker~\cite{tmt} attempts to enhance features of the search image by correlating them with the features of multiple historical templates.
TransT~\cite{transt} and Stark~\cite{stark} enhance features of both the template image and search image based on attention which is more powerful than the standard linear correlation operation in Siamese tracker~\cite{SiameseFC,siamrpn, siamrpn++}. The above methods outperform the previous state-of-the-art methods by a notable margin. Our work differs from ~\cite{tmt,transt,stark} in that we discard CNN and use pure Transformer to extract features.

\section{Dual-branch TransFormeR (DualTFR)}
This section presents the technical details of DualTFR. Section~\ref{sec:Overall} first gives an overview. 
Then we dive into the details of DualTFR including local/global attention and cross attention in section~\ref{sec:block}. In section~\ref{sec:variants}, we describe multiple variants of DualTFR .

\subsection{Architecture Overview}
\label{sec:Overall}
As in Fig.~\ref{fig:tfr}, there are two branches in DualTFR, one for the search image $x$ and the other for the template image $z$. Both are split into non-overlapping patches of equal size ($4\times4$ pixels), respectively. Each of the patches is treated as a token. In total, there are  $\frac{H}{4} \times \frac{W}{4}$ tokens, with each having a $48$ dimensional feature vector.

\paragraph{Transformer-based Feature Extraction}
We first apply a linear projection layer to increase the feature dimension from $48$ to $C$ for all tokens.
Then the resulting template feature maps $f_z \in \mathbb{R}^{4C \times \frac{H_z}{16} \times \frac{W_z}{16} }$
and search feature maps $f_x \in \mathbb{R}^{4C \times \frac{H_x}{16} \times \frac{W_x}{16} }$ are fed to Local Attention Blocks (LAB). The LAB weights are shared between the two branches. Note that LAB only computes attention within a small window with $7 \times 7$ tokens in order to reduce the computation time. 
A number of LAB and patch merging layers are stacked as shown in Fig.~\ref{fig:tfr}.

The patch merging layer is used to decrease the spatial resolution and increase the channel dimension of the feature maps both by a factor of two.
The resolutions of the template and search feature maps after the LAB stage are $f_z \in \mathbb{R}^{4C \times \frac{H_{z}}{16} \times \frac{W_{z}}{16} }$ and $f_x \in \mathbb{R}^{4C \times\frac{H_{x}}{16} \times \frac{W_{x}}{16} }$, respectively. 
Then the two feature maps are fed to two Global Attention Blocks (GAB), respectively. Different from LAB, GAB computes attention among all tokens of the same image which allows to capture long-range dependency. 
Finally, they go into the Cross Attention Block (CAB) which computes attention among tokens from both images. 
The final resolution of the search feature maps remains the same as input. In practice, we concatenate the output features from the last two layers. So the resolution of the output feature is $f_{\text{out}} \in \mathbb{R}^{8C \times \frac{H_{x}}{16} \times\frac{W_{x}}{16} }$. We feed them to the prediction head, which will be described in detail in the subsequent section, to estimate the target location and shape.       

\vspace{-2mm}
\paragraph{Prediction Head}
Similar to Siamese-based trackers~\cite{SiameseRPN, SiamFCpp}, we add a prediction head to the output features $f_{\text{out}}$ to estimate whether each token (location) contains the target object and its offset and size. The first is formulated as a binary classification task while the second as a continuous regression task. In particular, the size of the object is represented by normalized width and height as in DETR~\cite{DETR}. Both are realized by multi-layer perception (MLP) network which consists of three linear projection layers and ReLU layers, respectively.

\subsection{Local, Global and Cross Attention}
\label{sec:block}

\paragraph{Multi-Head Attention}
After the image is split into tokens, we use pure attention operators to extract features following ViT \cite{vit}. The Multi-Head Attention (MHA) is the core of the approach so we briefly describe it to make the paper self-contained. MHA takes its input in the form of three parameters, known as Query $\mathbf{Q}$, Key $\mathbf{K}$ and Value $\mathbf{V}$. The three parameters are similar in structure. MHA is computed as:
\begin{equation}
{\text{MHA}}(\mathbf{Q},\mathbf{K},\mathbf{V}) = {\text{Concat}}({\mathbf{H}_1},...,{\mathbf{H}_{{n_h}}}){\mathbf{W}^O}, 
\end{equation}
\begin{equation}
{\mathbf{H}_i} = {\text{Attention}}(\mathbf{Q}\mathbf{W}_i^Q,\mathbf{K}\mathbf{W}_i^K,
\mathbf{V}\mathbf{W}_i^V),
\end{equation}
where $\mathbf{W}_{i}^{Q} \in \mathbb{R}^{d_{m} \times d_{k}}, \mathbf{W}_{i}^{K} \in \mathbb{R}^{d_{m} \times d_{k}}, \mathbf{W}_{i}^{V} \in$
$\mathbb{R}^{d_{m} \times d_{v}}$, and $\mathbf{W}^{O} \in \mathbb{R}^{n_{h} d_{v} \times d_{m}}$ are learnable parameters. More details about attention can be referred to~\cite{vaswani2017attention}.

\begin{figure}[t]
	\centering{\includegraphics[scale = 0.42]{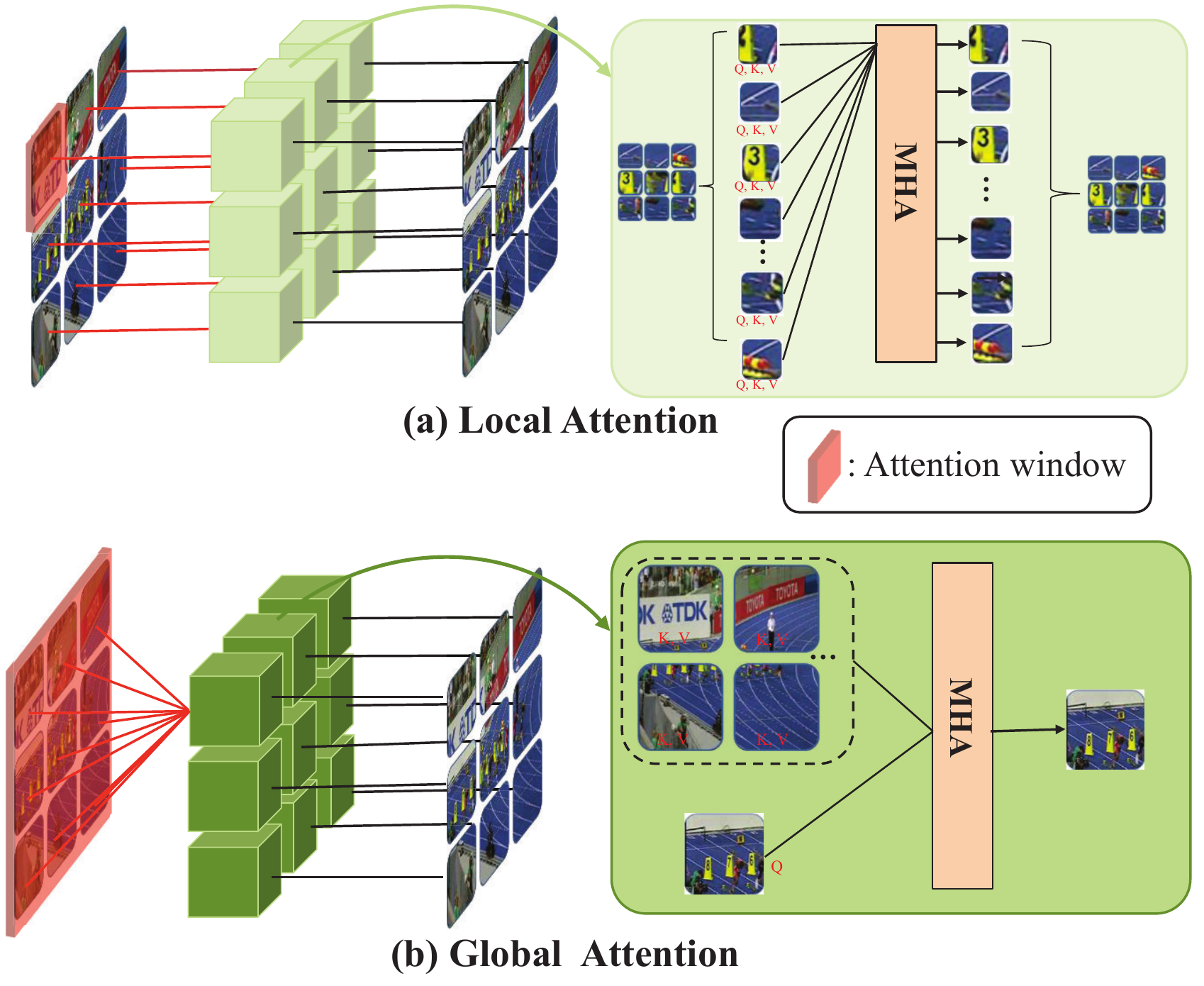}}
	\caption{Local Attention and Global Attention. MHA denotes multi-head attention. Attention computation only occurs inside the attention window.}
	\vspace{-3mm}
	\label{fig:attention}
\end{figure}

\paragraph{Local Attention}
The main difference between LAB and GAB lies in the size of the window to compute attention. For local attention, we only compute attention for tokens among a small window $M \times M$. In our experiments, $M$ is set to be $7$. Suppose an image is split into $N$ non-overlapping local windows and each window has $M \times M$ tokens. Then the computation cost is: 
\begin{equation}
F L O P_{Local}=4 N C^{2}+2 (M \times M)^{2} C ,
\end{equation}

\paragraph{Global Attention}
Global attention compute attention for all tokens in the same image. It has the capability to model long-range dependency across the whole image. But it also brings heavy computation burden. In specific, the computation cost is:
\begin{equation}
F L O P_{Global}=4 N C^{2}+2 M^{2} N C ,
\end{equation}
where $C$ is the number of channel dimension. 
The complexity of global attention is quadratic to the total number of tokens. In contrast, the complexity of the local attention is linear to the total number of tokens. Since the number of tokens $M^{2}$ within a small local window is fixed, the whole complexity is linear to the image size. As a result, we only use global attention when the resolution of the feature map is small. 

\begin{figure}[t]
	\centering{\includegraphics[scale = 0.45]{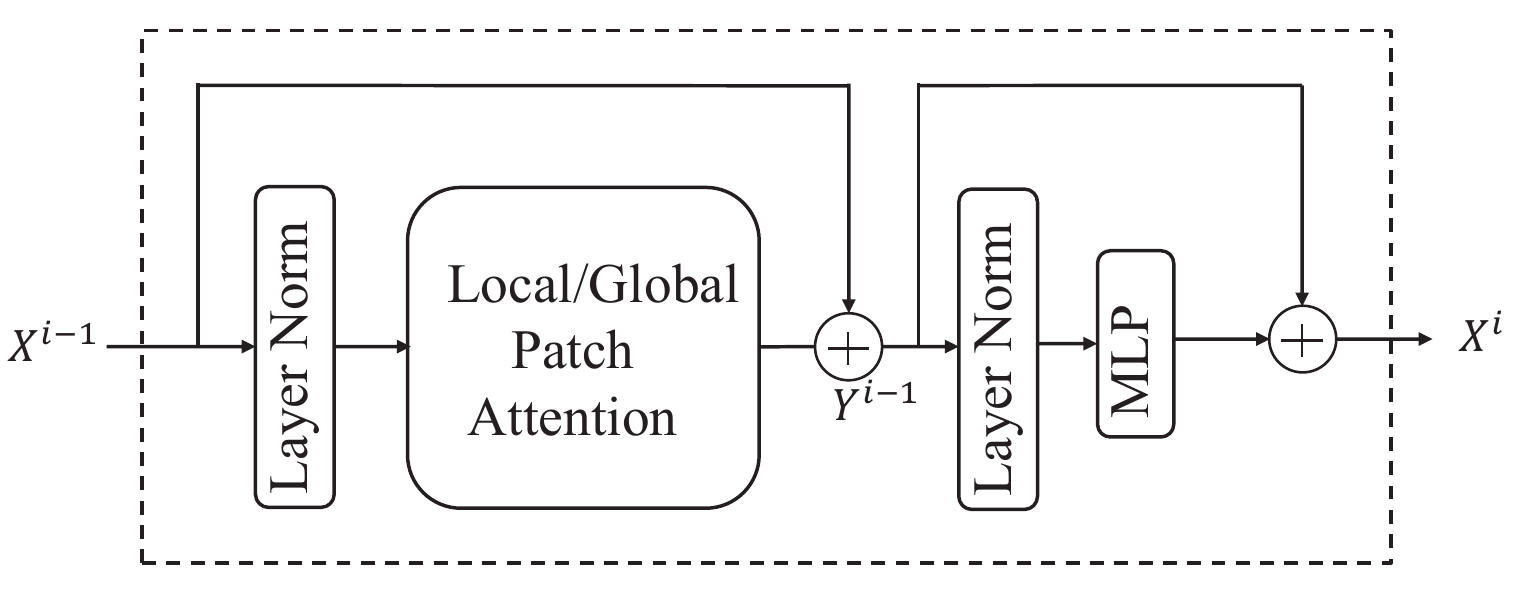}}
	\caption{The network structure of a Global or Local Attention Block. MLP denotes Multi-Layer Perception.}
		\vspace{-3mm}

	\label{fig:block}
\end{figure}

Fig.~\ref{fig:block} shows the structure of a GAB or LAB.
Layer normalization, multi-layer perception (MLP) and residual connection are used as in standard transformers. Mathematically, it is computed as:
\begin{equation}
\begin{aligned}
&\hat{\bf{Y}}^{i}=\text{MHSA}\left(\text{LN}\left(\bf{X}^{i-1}\right)\right)+\bf{X}^{i-1}, \\
&\bf{X}^{i}=\text{MLP}\left(\text{LN}\left(\hat{\bf{Y}}^{i}\right)\right)+\hat{\bf{Y}}^{i},
\end{aligned}
\end{equation}
where $\bf{X}^{i}$ denotes the output from the block $i$ and $\bf{Y}^{i}$ denotes the output from MLP in block $i$. $\text{MLP}$ represents multi-layer perception. $\text{MHSA}$ denotes multi-head self attention. In practice, we add the shifted-window mechanism following the~\cite{swin} for enhanced multi-scale feature representation.  

\paragraph{Cross Attention}
After we compute features for the template and search image, respectively, with LAB and GAB, we propose dual-branch Cross Attention to fuse features between the two images. It is similar to GAB except that we compute attention among tokens from both images. More specifically, in template branch, the template tokens are as key and value and search tokens are as query. Search branch is on the contrary. The operation allows to smooth appearance variations in neighboring frames which notably improves the tracking accuracy according to our experiments. Since we only use Cross Attention when the resolutions of the feature maps are small, the additional computation burden is not significant.

\subsection{Architecture Details}
\label{sec:variants}
In order to strike a good balance between tracking accuracy and speed, we evaluate multiple parameter choices. We choose the following hyper-parameters which achieves $40$fps inference speed on a single Nvidia 2080Ti GPU. We set the projection dimension
$C$ to be $128$. The window size in LAB is $7 \times 7$. Each token has $4 \times 4$ pixels. The numbers of layers in LA, GA, CA blocks are $10, 6, 4$, respectively. Details can be seen in Fig.~\ref{fig:tfr}.

\section{Details, Datasets and Metrics}
This section describes implementation details, datasets and metrics, and the results of the state-of-the-art methods. To validate the effectiveness of DualTFR, we all adopt fixed template strategy with no other tricks except for VOT2021 benchmark.

\subsection{Implementation Details} \label{sec:Implementation}
\paragraph{Training}
We train the model in two steps. In the first, we pre-train LAB on the large scale ImageNet-1K~\cite{ImageNet} dataset in the context of classification. The dataset contains $1.28M$ training images from 1000 classes. Following ~\cite{swin}, we employ an AdamW~\cite{AdamW} optimizer and train the model for $300$ epochs. The batch size is 512 and the learning rate is $10^{-5}$ with 0.05 weight decay.

Next, we finetune the whole model on the tracking datasets. In particular, for each pair of search/template images from the training dataset, we compute the losses based on the classification and regression outputs from the prediction head. We use standard cross-entropy loss for the classification loss: all pixels within the ground-truth box are regarded as positive samples and the rest are negative. We use GIoU~\cite{GIoULoss} loss and $L_1$ loss for the regression loss. We load the pre-trained LAB parameters and initialize the rest of the parameters by Xavier~\cite{xai}. We use $8$ tesla V100 GPUs and set the batch size to be $480$. The search area factor of template and search image is set to 1.5 and 3, respectively. The total sample pairs of each epoch is $40$ million. The learning rate is set to be $10^{-5}$ for the pre-trained weights, and $10^{-4}$ for the rest. The learning rate decays by a factor of $10$ at the $40_{th}$ epoch. We finetune the model for $100$ epochs. 

The training datasets include the train subsets of LaSOT~\cite{LaSOT}, GOT-10K~\cite{GOT10K}, COCO2017~\cite{COCO}, and TrackingNet~\cite{trackingnet}. All the forbidden sequences defined by the VOT2019 challenge are removed. The pairs of training images in each iteration are sampled from one video sequence. On static images, we also construct an image pair by applying data augmentation like flip, brightness jittering and target center jittering.   

\paragraph{Inference Details}
During inference, the regression head and classification head generate feature maps which contain estimated box shapes and location confidence values. The maximum confidence value and its corresponding bounding box size are chosen to be final prediction result. The template and search image size are set to $112 \times 112$ and $224 \times 224$, respectively.
We also evaluate our approach with two tricks on VOT2021 benchmark. The approach with the first trick is the spatio-temporal version. Inspired by \cite{stark}, we obtain a global context vector from the search branch feature in the previous frame by global average pooling and add it as a new token to the template token set. The update interval is set to one.
The second is the online version, an online correlation filter is added to the model. The response map from online filter is added to the classification map with the weight value of $0.2$. Note that the two tricks are complementary to our approach.

\begin{table*} [t]
	\begin{center}
		\tabcolsep=2pt
		\resizebox{2.0\columnwidth}{!}{ %
			\begin{tabular}{c c | c c c c c c c c c c}
				\toprule
				\multicolumn{2}{c}{Trackers} &
				\begin{tabular}{c} SiamFC++ \cite{SiamFCpp} \\  \end{tabular} &
				\begin{tabular}{c} SiamRPN++ 
					\cite{siamrpn++}\\  \end{tabular} &
				\begin{tabular}{c} ATOM\cite{atom} \\  \end{tabular} &
				\begin{tabular}{c} DiMP-50\cite{DiMP} \\   \end{tabular}  &
				\begin{tabular}{c} D3S\cite{D3S} \\ \end{tabular}  &
				\begin{tabular}{c} Ocean\cite{Ocean} \\  \end{tabular}  &
				\begin{tabular}{c} SAMN\cite{samn} \end{tabular} &
				\begin{tabular}{c}  STARK-S50~\cite{stark} \end{tabular} &
				\begin{tabular}{c} TransT\cite{transt} \end{tabular} &
				\begin{tabular}{c} Ours \end{tabular}\\
				
				\midrule
				\multirow{2}{*}{GOT-10K}

				& AO$\uparrow\ $
			    & 59.5
				& 51.8 & 55.6
				&61.1&59.7 & 61.1 & 
				61.5
				& {\color{green}67.2}
				& {\color{blue}72.3}
				& {\color{red}73.5}\\
				
				& SR\textsubscript{.50}$\uparrow\ $
				 & 69.5
				& 32.5 & 63.4
				& 71.7
				& 67.6 
				& 72.1
				& 69.7
			    &  {\color{green}76.1}
				& {\color{blue}83.7}
				& {\color{red}84.8}\\
				
				& SR\textsubscript{.75}$\uparrow\ $
			    & 47.9
				& 61.6 & 40.2
				& 49.2
				& 46.2
				& 47.3
				& 52.2
				& {\color{green}61.2}
				& {\color{blue}68.1} 		
				& {\color{red}69.9} \\
				
				\bottomrule
			\end{tabular}%
		}
		\end{center}
		\vspace{-5mm}
		\caption{Results on GOT-10K. Top-3 results of each dimension (row) are colored in red, blue and green, respectively.}
				\vspace{-3mm}
		\label{tab:got10k}
	\end{table*}
	
\begin{table*} [!h]
		\begin{center}
			\tabcolsep=2pt
			\resizebox{2.0\columnwidth}{!}{ %
				\begin{tabular}{c c|  c c c c c c c c c c c}
					\toprule
					\multicolumn{2}{c}{Trackers} &
					\begin{tabular}{c} SiamRPN++ \cite{siamrpn++}\\  \end{tabular} &
					\begin{tabular}{c} SiamFC++ \\  \end{tabular} &
					\begin{tabular}{c} DiMP50 \\  \end{tabular} &
					\begin{tabular}{c} MAML-FCOS \cite{MAML-track}\\  \end{tabular}&
					\begin{tabular}{c} SAMN \cite{samn}\\  \end{tabular} &
					\begin{tabular}{c} SiamAttn \cite{SASiam}\\  \end{tabular} &
					\begin{tabular}{c} PrDiMP50\cite{PrDiMP} \\  \end{tabular}&
					\begin{tabular}{c} STARK-S50~\cite{stark} \\  \end{tabular}&
					\begin{tabular}{c} Ours \end{tabular} \\
					
					\midrule
					\multirow{3}{*}{TrackingNet}
					
					& AUC(\%)$\uparrow\ $
					& 73.3 & 75.4
					& 74.0 & {\color{green}75.7}
					&  74.2 & 75.2
					& {\color{green}75.8}
     				& {\color{red}80.3}
					&  {\color{blue}80.1} \\
					
					& P$_{norm}$(\%)$\uparrow\ $
					& 80.0 & 80.0
					& 80.1 & 82.2
					& 79.4 & {\color{green}81.7}
					& 81.6
					&{\color{red}85.1}
					& {\color{blue}84.9}   \\
					
					\bottomrule
				\end{tabular}%
			}
			\end{center}
		\vspace{-5mm}
			\caption{Results on TrackingNet.}
			\label{tab:tknet}
					\vspace{-3mm}
		\end{table*}
		
\begin{table*} [!h]
			\begin{center}

				\tabcolsep=2pt
				\resizebox{2.0\columnwidth}{!}{ %
					\begin{tabular}{c c| c c c c c c c c c c c}
						\toprule
						\multicolumn{2}{c}{Trackers} &
						
			 	   \begin{tabular}{c} MDnet\cite{MDNet} \\  \end{tabular} &
						\begin{tabular}{c} ECO\cite{ECO}\\  \end{tabular} &
						\begin{tabular}{c} ATOM \\  \end{tabular} &
						\begin{tabular}{c} SiamBAN~\cite{siamban}\\  \end{tabular} &
						\begin{tabular}{c} SiamCAR\cite{SiamCAR} \\  \end{tabular} &
						\begin{tabular}{c} MAML~\cite{MAML-track} \\  \end{tabular} &
						\begin{tabular}{c} PrDiMP \cite{Kristan2019a}\\ \end{tabular}  &
						\begin{tabular}{c} SiamFC++\\   \end{tabular}  &
						\begin{tabular}{c} Ocean~\cite{Ocean} \\ \end{tabular}  &
						\begin{tabular}{c} TransT~\cite{transt} \\ \end{tabular}  &
						\begin{tabular}{c} Ours \end{tabular} \\

						\midrule
						\multirow{3}{*}{LaSOT}
						& AUC
						$\uparrow\ $
					    & 39.7
						& 32.4 & 51.5
						& 51.4 & 50.7
						& 52.3 & {\color{green}59.8}
						&54.4 & 56.0&{\color{red}64.9} & {\color{blue}63.5}\\
						
						& P$_{norm}\uparrow\ $
						 & 46.0
						&33.8 & 57.6
						& 59.8 & 60.0
						& - & {\color{blue}68.8} &62.3
						& - & {\color{red}73.8} & {\color{blue}72.0} \\
						
						& P$\uparrow\ $
						 & 37.3
						& 30.1 & 50.5
						& 52.1 & 51.0
						& - & {\color{green}60.8}
						& 54.7 & 56.6
						& {\color{red}69.0} & {\color{blue}66.5}\\
						
						\bottomrule
				\end{tabular}}%
			\end{center}
		\vspace{-5mm}
			\caption{Results on LaSOT.}
			\label{tab:lasot}
					\vspace{-3mm}
		\end{table*}
		
\begin{table*} [!h]
			\begin{center}

				\tabcolsep=2pt
				\resizebox{2.0\columnwidth}{!}{ %
					\begin{tabular}{c c|  c c c c c c c c c c c}
						\toprule
						\multicolumn{2}{c}{Trackers} &
						\begin{tabular}{c} DPMT\cite{xie2020hierarchical} \\  \end{tabular} &
						\begin{tabular}{c} SuperDiMP~\cite{SuperDiMP} \cite{sdimp} \\  \end{tabular} &
						\begin{tabular}{c} DiMP \\  \end{tabular} &
						\begin{tabular}{c} ATOM \\  \end{tabular} &
						\begin{tabular}{c} SiamMask~\cite{SiamMask} \\  \end{tabular} &
						\begin{tabular}{c} STM~\cite{STM} \\  \end{tabular} &
						\begin{tabular}{c} DET50~\cite{Kristan2020a} \\ \end{tabular}  &
						\begin{tabular}{c} Ocean\\   \end{tabular}  &
						\begin{tabular}{c} TransT\cite{transt} \\
						\end{tabular}&
						\begin{tabular}{c} Stark-S50~\cite{stark} \\
						\end{tabular}&
						\begin{tabular}{c} Ours 
						\end{tabular} \\
						
						\midrule
						\multirow{3}{*}{VOT-20}
						
						& Acc.$\uparrow\ $
						& 0.492 & 0.492
						& 0.457 & 0.462
						& 0.624
						&  {\color{green}0.751} & 0.679
						& 0.693 & -
						& {\color{red}0.761}
						& {\color{blue}0.755}  \\
						
						& Rob.$\uparrow\ $
						& 0.745 & 0.745
						& 0.740 & 0.734
						& 0.648
						& 0.574 & {\color{green}0.787}
						& 0.754 & - 
						& {\color{blue}0.749 }
						& {\color{red}0.836}  \\
						
						& EAO$\uparrow\ $
						& 0.303 & 0.305
						& 0.274 & 0.271
						& 0.321
						& 0.308 & 0.441
						& 0.430 & {\color{blue}0.495} 
						& {\color{green}0.462}
						& {\color{red}0.528} \\
						
						\bottomrule
				\end{tabular}}%
			\end{center}
		\vspace{-5mm}

			\caption{Results on VOT2020. We use AlphaRefine\cite{AlphaRefine} to generate mask for VOT benchmark.}
			\label{tab:vot20}
		\vspace{-3mm}

		\end{table*}

\begin{table} [t]
	\begin{center}
		\tabcolsep=1.5pt
		\resizebox{1\columnwidth}{!}{ %
			\begin{tabular}{c c| c c c c c c c c c }
				\toprule
	   \multicolumn{1}{c}{} &
		  	 \begin{tabular}{c} \\  \end{tabular} &
	    	\begin{tabular}{c} ATOM \\  \end{tabular} &
	    	\begin{tabular}{c} SiamRPN++ \\  \end{tabular} &
	    	\begin{tabular}{c} DiMP \\  \end{tabular} &
	        \begin{tabular}{c} STMTrack~\cite{stmtrack} \\  \end{tabular} &
	    	\begin{tabular}{c} SiamRN~\cite{saimrn} \\  
	    	\end{tabular} &
			Ours 
			\\
				\midrule
				& AUC
				& 64.3
				& 61.3 
				& {\color{green}65.3}
				& 64.7
				& {\color{blue}64.8} 
				& {\color{red}68.2}
				\\
				
				\bottomrule
		\end{tabular}}%
	\end{center}
		\vspace{-5mm}

	\caption{Results on UAV123.}
			\vspace{-3mm}

	\label{tab:uav}

\end{table}

\begin{table} [t]
	\begin{center}
		\tabcolsep=2pt
		\resizebox{0.8\columnwidth}{!}{ %
			\begin{tabular}{c c|  c c c | c c c }
				\toprule
				\multicolumn{1}{c}{} &
		  	 \begin{tabular}{c}  \\  \end{tabular} &
		  	 
	    	\begin{tabular}{c}  \\  \end{tabular} &
			Baseline
			\begin{tabular}{c}  \\  \end{tabular} &  
			\begin{tabular}{c}  \\  \end{tabular} &
			\begin{tabular}{c}  \\  \end{tabular} &
			
			\begin{tabular}{c}  Realtime \\ \end{tabular} &
			\\
	   \multicolumn{1}{c}{} &
		  	 \begin{tabular}{c} \\  \end{tabular} &
	    	\begin{tabular}{c} EAO \\  \end{tabular} &
	    	\begin{tabular}{c} Acc. \\  \end{tabular} &
	    	\begin{tabular}{c} Rob. \\  \end{tabular} &
	    	\begin{tabular}{c} EAO \\  \end{tabular} &
	    	\begin{tabular}{c} Acc. \\  \end{tabular} &
			Rob. \\
				\midrule
				& Baseline
				& 0.525 & 0.748
				& 0.826
				
				& 0.509 & 0.746
				& 0.815\\
				& ST
				& 0.536 & 0.755
				& 0.836
				& {\color{red}0.512} & {\color{red}0.751}
				& {\color{red}0.816}\\
				& On
				& {\color{red}0.539}  & {\color{red}0.757} 
				& {\color{red}0.837} 
				
				& 0.395 & 0.681
				& 0.741\\
				
				\bottomrule
		\end{tabular}}%
	\end{center}
		\vspace{-5mm}

	\caption{Results on VOT2021. ST denotes the spatio-temporal version of DualTFR. ``On'' denotes the online version of DualTFR.}
	\label{tab:vot21}
		\vspace{-5mm}

\end{table}

\subsection{Evaluation}

We compare DualTFR to the state-of-the-art trackers on five tracking benchmarks. Moreover, we also report results on the recently introduced VOT2021 benchmark.  

\begin{figure}[t]
	\centering{\includegraphics[scale = 0.45]{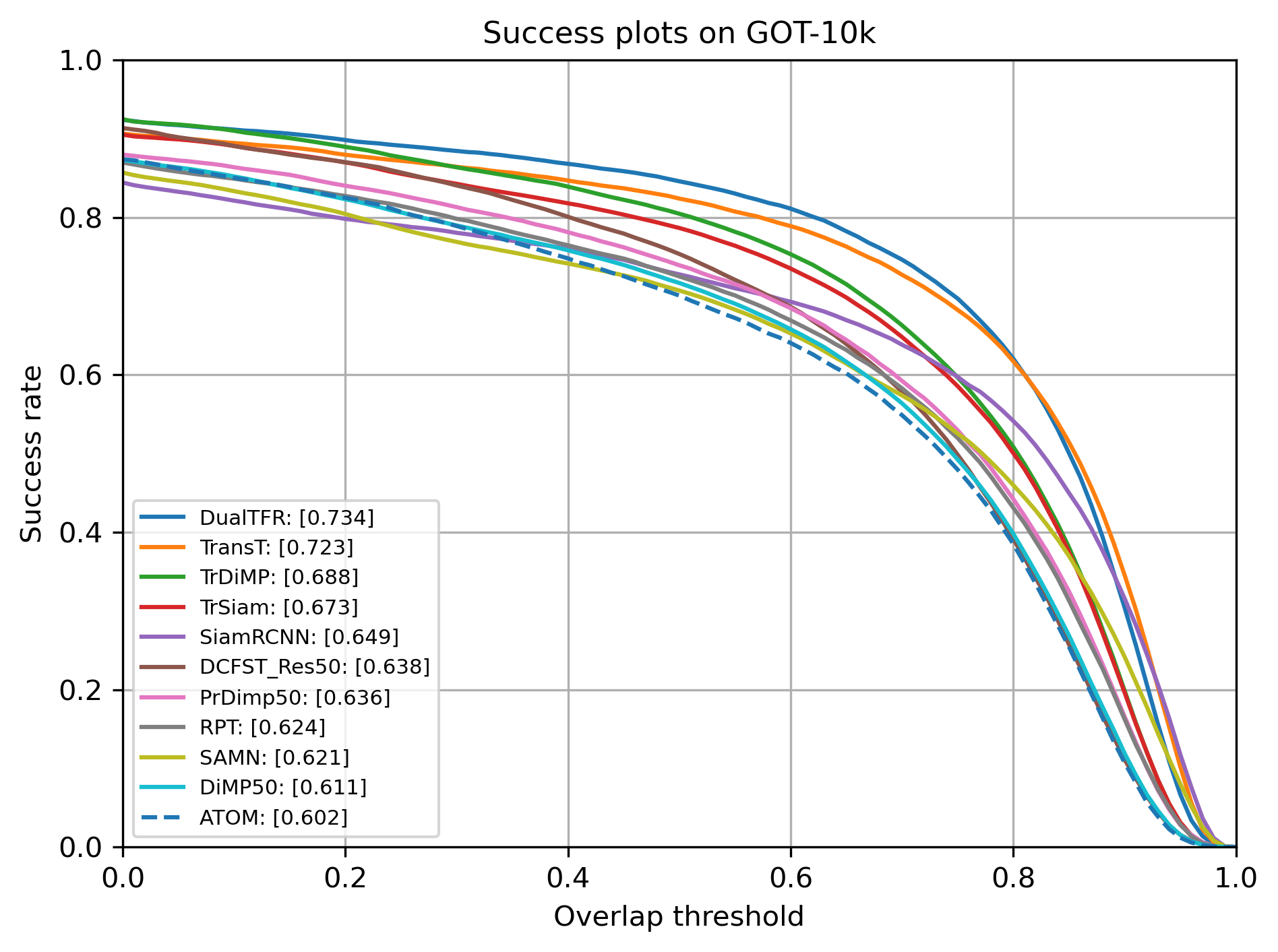}}
    \vspace{-3mm}
	\caption{Comparisons on GOT-10k test set. }
	\vspace{-5mm}

	\label{fig:got}

\end{figure}

\vspace{-3mm}
\paragraph{GOT-10K} We only compare to the trackers which use additional training datasets for fair comparison. Results are obtained from the official evaluation server. As shown in Table \ref{tab:got10k} and Fig.~\ref{fig:got}, our tracker outperforms all competing trackers in terms of three metrics and achieves the best $AO$ score of $73.5$. We also compare to a transformer-based tracker TransT~\cite{transt}, our tracker improves the $SR_{75}$ by 1.8 points while raises the $AO$ by 1.2 points.
As for the fully CNN-based Siamese correlation trackers, DualTFR outperforms Ocean~\cite{Ocean} by $12.4$ points in terms of $AO$. The results validate the values of using Transformers to extract features.

\vspace{-3mm}
\paragraph{TrackingNet} TrackingNet contains $511$ test video sequences. We report the Success (AUC) and Precision ($P_{norm}$) results in Table \ref{tab:tknet}, DualTFR achieves comparable results with STARK-S50~\cite{stark}. Please note that DualTFR adopts $224 \times 224$ as search image size which is smaller than $320 \times 320$ in STARK. Both network stride of DualTFR and STARK is 16. Thus, we claim that the discriminative ability in smaller image size of DualTFR is more powerful.  SiamAttn~\cite{Deform_siam} is a Siamese tracker with attention generated from convolution operation.
DualTFR improves SiamAttn by $4.9$ points in terms of AUC and $3.2$ points in precision.

\vspace{-3mm}
\paragraph{LaSOT} 
LaSOT contains $280$ long-term video sequences for testing. The evaluation protocol we adopted is one-pass evaluation. The success rate (AUC) and precision (P) of recent sota trackers are presented in Table~\ref{tab:lasot}. DualTFR achieves comparable results with TransT~\cite{transt} and surpasses remaining trackers in three metrics. The main reason that DualTFR does not perform better than TransT lies in the network stride. TransT adopts stride 8 and $32 \times 32$ output size while DualTFR adopts stride 16 and $14 \times 14$ output size. Smaller stride is preferred as addressed in~\cite{siamdw}. DualTFR can be improved with smaller stride in the future.

\vspace{-3mm}
\paragraph{UAV123} UAV123 contains 123 aerial video sequences of small objects captured from low latitude UAVs. 
One pass evaluation protocol is adopted (AUC denotes success rate). Table.~\ref{tab:uav} shows the results. Compared to the recent SOTA trackers SiamRN\cite{saimrn} and STMTrack\cite{stmtrack},  DualTFR has over $3.4$ improvements on AUC and achieves better performance than the remaining trackers.

\vspace{-3mm}
\paragraph{VOT2020} 
VOT2020 adopts an anchor-based evaluation protocol which conducts multiple tests for one video sequence without reset operation. VOT2020 accepts axis-aligned, rotated box or binary segmentation mask format. The final metric for ranking is the Expected Average Overlap (EAO). Here, we use the alpha-refine~\cite{AlphaRefine} for mask generation. 
The VOT2020 top performers are RPT~\cite{RPT} and Ocean~\cite{Ocean}, two recent sota transformer-based trackers Stark~\cite{stark} and TransT~\cite{transt}, and classical deep trackers SiamRPN++~\cite{siamrpn++}, ATOM~\cite{atom}, DiMP~\cite{DiMP} and sota video obeject segmentation method STM~\cite{STM} are compared with our tracker.
Table.~\ref{tab:vot20} shows that our tracker outperforms all trackers on all three measures. In terms of EAO, DualTFR outperforms the strongest SOTA Stark-50 by $2.3$ points and TransT by $3.3$ points. DualTFR has larger improvement over other methods. Note that transformer-based trackers (Stark, TransT, DualTFR) are nearly or higher than $0.50$ EAO, it shows the superiority of  attention-based models towards current fully CNN-based trackers.

\vspace{-3mm}
\paragraph{VOT2021} The evaluation metrics and protocol on the VOT2021 dataset is the same with VOT2020 benchmark. A certain number of hard video sequences are chosen to replace the easy video sequences on VOT2020. As described in Sec.~\ref{sec:Implementation}, we presents three versions of DualTFR in Table.~\ref{tab:vot21} to show its wide applicability. DualTFR achieves very competing performance on the VOT2021 benchmark.

\begin{figure*}[t]
	\centering{\includegraphics[scale = 0.57]{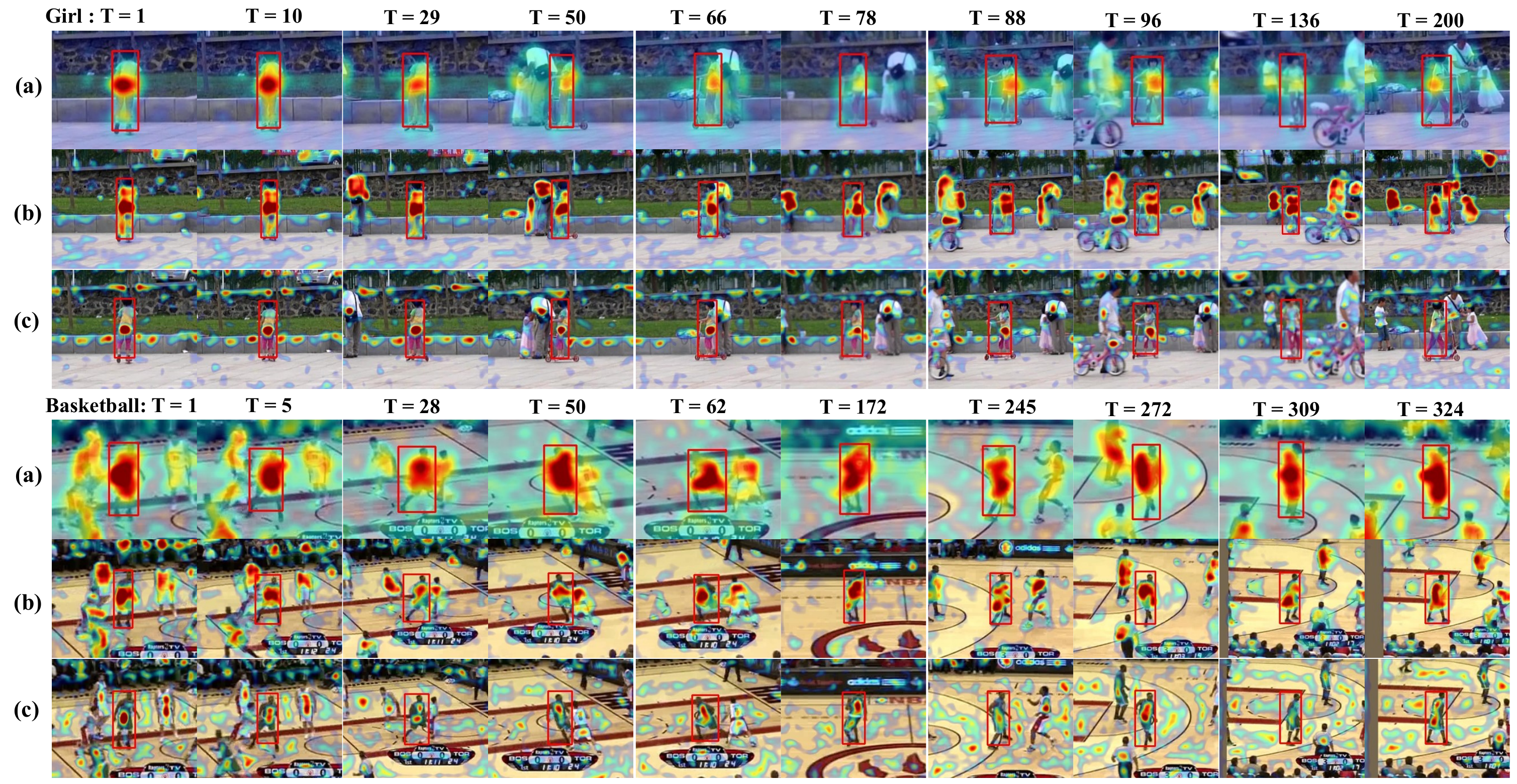}}
	\vspace{-5mm}
	\caption{Visualization results. The cosine similarities between center point of template and the whole search-region feature. Features are from the the last layer of CNN backbone or the last block of LAB. (a) (b) (c) are referred to the results of DualTFR, TransT, SiamRPN++. Note that the search area scale of DualTFR is smaller, but it does not influence our analysis. }
	\label{fig:v1}
	\vspace{-3mm}

\end{figure*}

\section{Ablation Studies}
In this section, we discuss the potential of the fully attention-based model in visual object tracking by a number of ablation studies. 

\subsection{Transformer Features vs. CNN Features}
To investigate why transformer-based features are better than the CNN-based features, we visualize the attention maps of the template and the search images before cross attention block. We select three trackers which also follow the Siamese framework for comparison. DualTFR belongs to the fully attention-based tracking. TransT is a representative method which combines CNN-based feature extraction and attention-based fusion, SiamRPN++ is a pure CNN-based Siamese tracker.

\vspace{-3mm}
\paragraph{Instance-Discriminative Features}
As shown in Fig.~\ref{fig:v1}, the visualization results of the CNN-based trackers (last two rows) all have large responses on all instances having similar appearance. See the two human instances in $245_{th}$ frame in the basketball example.
Note that the response map indicates the similarity between the feature of the template center point and all features from the search image. 
As for the attention-based features, the similarities between template center and distractor objects are much lower which greatly enhances the discriminative ability of tracking model.
Another interesting phenomenon is that the high responses inside the target instance gradually expands from pure CNN-based, CNN+transformer to fully Transformer-based model. 
For the pure CNN-based tracker SiamRPN++, the response values are very focus and narrow which means the template center only shares high similarity with the exact center point of target instance. 
In contrast, the response map of DualTFR has high values almost over the whole target area. 
It indicates that attention-based feature network is more focus on inter-instance difference rather than intra-instance. We name it instance-discriminative features.
Thus, we argue that the attention-based features learned from matching are more suitable for instance-level tasks.
In Table.~\ref{tab:ab1}, we replace the LAB and GAB in DualTFR by ResNet-50 which has comparable parameters. The performance on GOT-10k drops $2.7$ points in terms of $AO$ from $73.5\%$ to $70.8\%$. This validates the superiority of attention-based feature. 

\begin{figure}[t]
	\centering{\includegraphics[scale = 0.32]{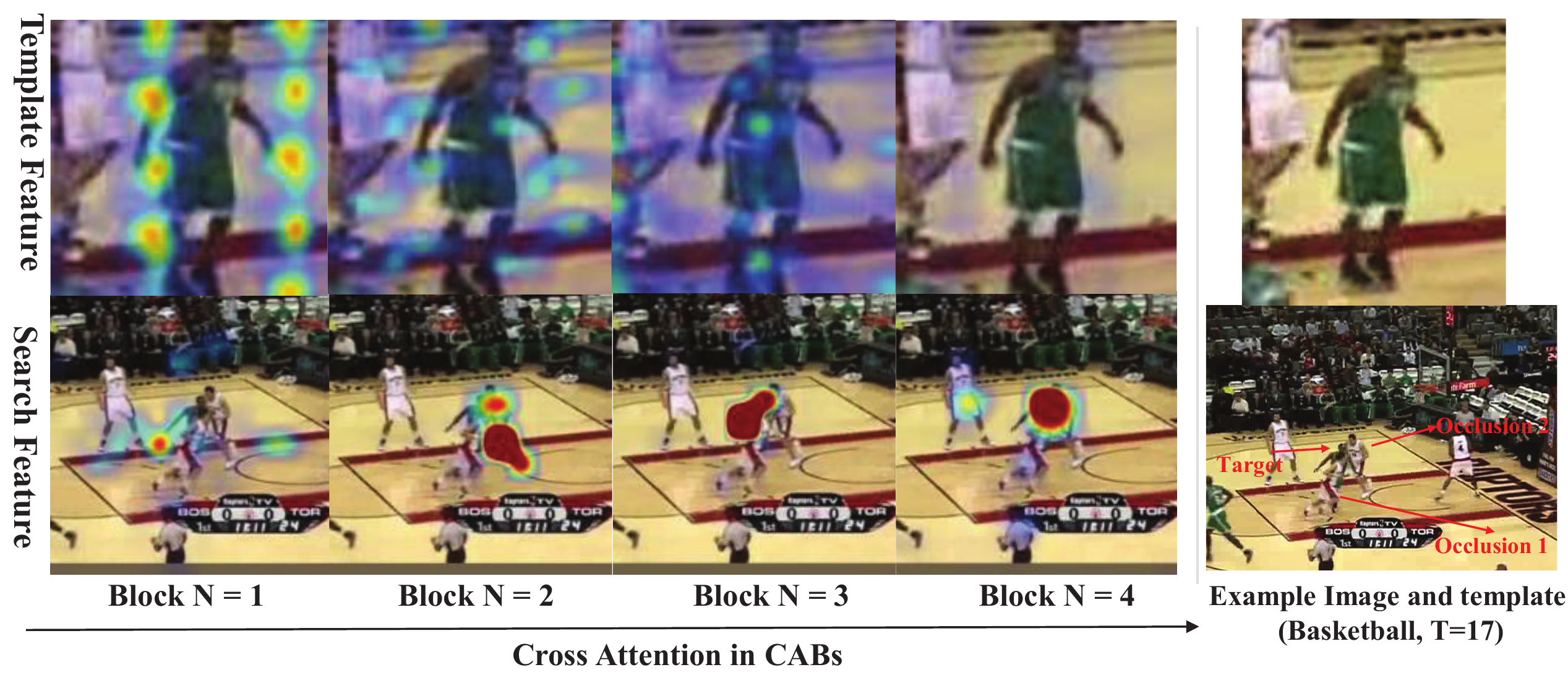}}
		\vspace{-4mm}
	\caption{Cross attention in template and search-region feature when the CAB goes deeper.}
	\label{fig:v_fusion}
	\vspace{-3mm}

\end{figure}

\vspace{-3mm}
\paragraph{Attention-Based Progressive Fusion Manner}
As illustrated in TransT~\cite{transt}, the transformer-based fusion performs better than linear convolution operation. Here, we further stress that the progressive manner of attention-based model are the main distinctions towards CNN-based pipelines. The attention-based CABs can gradually exchange the information between template and search branch which allows for the progressive refinement. 
As shown in Fig.~\ref{fig:v_fusion}, the attention in search-region gradually focuses on the target and distinct the distractors on the nearby spatial location. 
In the meantime, the feature embedding of template is adaptive to the search-region feature . More specifically, the template feature refines itself to be a more abundant feature bank for matching the search-region feature.
In Table.~\ref{tab:ab1}, we replace the CABs with depth-wise correlation which results in a direct drop of performance from $73.5\%$ to $51.2\%$ in GOT-10k. This demonstrates the advantage of attention-based fusion. 

\subsection{ImageNet Pre-training Vs. Train from Scratch}
For traditional deep trackers, the backbone with ImageNet pretrained parameters is vital for learning tracking representations. As shown in Table.~\ref{tab:ab1}, if the weight-sharing LAB part is pre-trained in ImageNet dataset, the whole performance rises from $48.5\%$ to $73.5\%$. This indicates that the fully transformer-based pipeline needs the prior knowledge from ImageNet pretraining. However, it brings extra training burden which we hope to bridge the gap between training from scratch and ImageNet-pretraining.

\begin{table}[t]
\centering
\small
\addtolength{\tabcolsep}{1.pt}
\begin{tabular}{c|ccccc}
\Xhline{1.0pt}

method & Backbone & Param. & FLOPs & EAO \\
\hline
SiamRPN++ & ResNet-50 & 53.9M & 59.5G & 0.356 \\
\hline
STARK-ST101  & ResNet-101 & 42.4M & 18.5 G & 0.497 \\
TransT  & ResNet-50 & 23M & 19.1G & 0.495 \\
\hline
DualTFR & LAB & 44.1M & 18.9G & 0.528 \\

\hline
\Xhline{1.0pt}
\end{tabular}
\normalsize
\caption{Comparison of parameters and flops. EAO denotes the performance on VOT2020 benchmark. The EAO of SiamRPN++ comes from the SiamMargin in~\cite{Kristan2020a}.}
\vspace{-1mm}

\label{tab:flop}
\end{table}

\begin{table}[t]
\small
\addtolength{\tabcolsep}{1.pt}\begin{center}

\begin{tabular}{c | c c c c}

\Xhline{1pt}
ImageNet.Pre       &\checkmark	&\xmark	&\checkmark	&\checkmark	\\
\hline
Trans.Extraction    &\checkmark	&\checkmark	& \xmark	&\checkmark\\

Conv.Extraction  &\xmark	&\xmark	&\checkmark	&\xmark \\
\hline

Trans.Fusion  &\checkmark	&\checkmark	&\checkmark	& \xmark	\\
Conv.Fusion  &\xmark	&\xmark	&\xmark	&\checkmark	\\
\hline
$AO$ $\uparrow\ $  &73.5	&48.5	&70.8	&54.2 \\
\Xhline{1pt}
\end{tabular}

\end{center}
\vspace{-3mm}

\caption{Ablation study on GOT-10k. \checkmark denotes the choice between two modules. \xmark  represents model does not choose this module. Trans.Fusion denotes transformer-based feature fusion. Conv.Fusion denotes depth-wise correlation in Siamese tracker. Extraction denotes the type of backbone. Trans. denotes transformer-based while Conv. denotes CNN-based.}
\vspace{-8mm}
\label{tab:ab1}

\end{table}

\subsection{Impacts of LAB, GAB and CAB}
We further investigate different configurations of blocks. For convenience and fair comparison, the ablation settings are all train-from-scratch and follow GOT-10k training protocol. With 2 GABs connected to the LAB, the performance of $AO$ rises from $42.1$ to $46.1$ comparing to no GAB settings. This is mainly because the global modelling enhances the feature representation of LAB. More stacked CABs which means more comprehensive feature fusion also brings improvements ($42.1$ to $46.1$). Though more LABs can provide better feature extraction ability, the parameters and flops increases sharply. With $4.7$ improvements in $AO$, 9 LABs brings extra 15M parameters and 12G flops. For a trade-off, we choose the fouth settings which has comparable parameters and flops to STARK to implement our DualTFR.

\begin{table}[t]
\small
\addtolength{\tabcolsep}{1.pt}\begin{center}

\begin{tabular}{c | c c c c c}

\Xhline{1pt}
LAB  &(2,2,6)	&(2,2,6)	&(2,2,6)  &(2,2,6) 	& (2,2,18)	\\

GAB  &\xmark	&\xmark    	& 2      & 4	    & 4\\

CAB   & 4	   & 2	       & 4	    & 4    & 4	 \\
\hline
Param.  & 34.1 M	& 29.2 M	 & 39.2 M	& 44.1 M  & 72.4 M	 \\
FLOPs.  & 14.2 M	& 11.3 G	 & 16.5 G	& 18.9 G  & 30.0 G	 \\

\hline

$AO$ $\uparrow\ $  &42.1	&40.4	&46.1	& 48.5   & 53.2 	 \\
\Xhline{1pt}
\end{tabular}
\end{center}
\vspace{-3mm}
\caption{Ablation study of LAB, GAB, and CAB in GOT-10k. The triplets of LAB denotes the number of blocks in 3 different stages ( See Fig.~\ref{fig:tfr} ). \xmark denotes not used.}
\vspace{-5mm}
\label{tab:ab1}
\end{table}

\subsection{Future Work}
As shown in Table~\ref{tab:flop}, DualTFR has comparable parameters and flops while outperforms the STARK in VOT2020 benchmark. 
It is worth noting that the parameters of DualTFR is twice of the TransT ($44.1M$ vs. $23M$). 
It is mainly due to the redundant independent module design. However, Siamese-style design may results in large computation flops. 
Inspired by ViT design manner, a unified dual-path block which can be stacked to formulate the whole tracking model may reduce the parameters and flops. In the future, DualTFR can be implemented by a more integral block.

\section{Conclusion}
 In this work, we propose a dual-branch fully transformer-based tracking architecture.   
 Through the dedicated design of GAB, LAB and CAB modules, we achieve a good balance between the computation cost and tracking performance. 
 Furthermore, we prove the superiority of fully attention-based paradigm to the traditional CNN-based tracking paradigm. 
In the future, the fully transformer-based tracking model can further be more light-weight through dedicated block design.
 Extensive experiments show that DualTFR performs at the state-of-the-art level while running at a real-time speed. 
 We hope this work could provides some insights on developing more powerful fully transformer-based trackers. 
 %

\section*{Acknowledgment}
We would like to thanks for advices from Chenyan Wu. This work was supported by NSFC (No.61773117 and No.62006041).%
{\small
	\bibliographystyle{ieee_fullname}
	\bibliography{ref}
}

\end{document}